\documentclass{article}
\usepackage{arxiv}
\usepackage[T1]{fontenc}

\usepackage{graphicx, caption,subcaption,ragged2e}
\usepackage{cite}
\usepackage{amsmath,amssymb,amsfonts}
\usepackage{algorithmic}
\usepackage{textcomp}
\usepackage{soul}
\usepackage{multirow}
\usepackage{hyperref}
\usepackage{xcolor}
\usepackage{pgfplots}
\usepackage{pifont}
\usepackage{arydshln}
\newcommand{\cmark}{\textcolor{green!80!black}{\ding{51}}}
\newcommand{\xmark}{\textcolor{red}{\ding{55}}}
\mathchardef\mhyphen="2D

\newcommand{\ourname}{\texttt{ConstScene}}

\begin{document}
\title{$\ourname$: A Dataset and Model for Advancing Robust Semantic Segmentation in Construction Environments}

\author{{Maghsood Salimi} \\
	School of Innovation, Design and Engineering, M{\"a}lardalen~University, Sweden  \\
	\texttt{maghsood.salimi@mdu.se} \\
\And
        \hspace{1mm}Mohammad~Loni \\
        Future Solutions Department, Volvo Construction Equipment, Sweden \\
	\texttt{mohammad.loni@volvo.com} \\
\AND
        \hspace{1mm}Sara Afshar \\
         Future Solutions Department, Volvo Construction Equipment, Sweden  \\
	\texttt{sara.afshar@volvo.com} \\
\AND
    \hspace{1mm}Antonio Cicchetti \\
         School of Innovation, Design and Engineering, M{\"a}lardalen~University, Sweden  \\
	\texttt{antonio.cicchetti@mdu.se}
 \AND
        \hspace{1mm}Marjan Sirjani \\
         School of Innovation, Design and Engineering, M{\"a}lardalen~University, Sweden  \\
	\texttt{marjan.sirjani@mdu.se}
}

\renewcommand{\shorttitle}{$\ourname$: A Dataset and Model for Advancing Robust Semantic Segmentation in Construction Environments}
\date{}

\maketitle              

\begin{abstract}

The increasing demand for autonomous machines in construction environments necessitates the development of robust object detection algorithms that can perform effectively across various weather and environmental conditions. This paper introduces a new semantic segmentation dataset specifically tailored for construction sites, taking into account the diverse challenges posed by adverse weather and environmental conditions. The dataset is designed to enhance the training and evaluation of object detection models, fostering their adaptability and reliability in real-world construction applications. Our dataset comprises annotated images captured under a wide range of different weather conditions, including but not limited to sunny days, rainy periods, foggy atmospheres, and low-light situations. Additionally, environmental factors such as the existence of dirt/mud on the camera lens are integrated into the dataset through actual captures and synthetic generation to simulate the complex conditions prevalent in construction sites. We also generate synthetic images of the annotations including precise semantic segmentation masks for various objects commonly found in construction environments, such as wheel loader machines, personnel, cars, and structural elements. To demonstrate the dataset's utility, we evaluate state-of-the-art object detection algorithms on our proposed benchmark. The results highlight the dataset's success in adversarial training models across diverse conditions, showcasing its efficacy compared to existing datasets that lack such environmental variability.

\keywords{Dataset \and Robust Object Detection \and Adversarial Attacks \and Semantic Segmentation \and Construction Environment}
\end{abstract}

\section{Introduction}
\label{sec:introduction}

The growing need for autonomous machines in construction environments underscores the importance of advancing resilient object detection algorithms \cite{somua2019computer, nath2020deep, kim2015robust}.
To ensure the applicability of the object detection model to unforeseen situations, the training dataset should include diverse and representative data encompassing various weather and lighting conditions \cite{ahmed2021survey, meng2022survey, sun2020scalability}.
Meeting this demand is crucial for enhancing the reliability of autonomous driving in the heavy machinery industry.

However, the scarcity of sufficient training data including diverse environmental conditions from construction environments hinders the ability of object detection models to generalize effectively.
In addition, it can exacerbate issues like overfitting, where models become overly specialized to the limited data available.
Addressing the problem of inadequate training data is pivotal for developing robust and versatile object detection models capable of meeting the demands of complex and dynamic environments.

To address these challenges, this paper introduces a large-scale semantic segmentation dataset designed exclusively for construction sites, considering the varied challenges presented by diverse weather conditions and situations occurring during construction operations.
We call this dataset robust semantic segmentation in construction environments, or in short: $\ourname$.
This dataset is designed to enhance the training and evaluation of semantic segmentation models, fostering their adaptability and reliability in real-world construction applications.

\textbf{Contributions.} $\ourname$ comprises annotated images captured under a wide range of weather conditions, including but not limited to sunny days, rainy periods, foggy atmospheres, and low-light situations.
Furthermore, environmental factors such as dust, and the existence of dirt or mud on the camera lens are systematically integrated into the dataset to simulate the complex conditions prevalent in construction environments.
The annotations include precise semantic segmentation masks for objects commonly found in construction environments, such as wheel loaders, personnel, cars, and structural elements.
We augment our dataset by (i) applying the blurring effect; (ii) adding noise; and (iii) removing random regions through a cut-out process simulating the presence of dirt or mud on the camera lens.
Finally, employing a monocular depth estimation model, we provide a depth map for every image in our dataset (Section~\ref{sec:data_collection:depth}). 

To demonstrate the utility of our dataset, we evaluate two state-of-the-practice and state-of-the-art object detection algorithms, U-Net \cite{ronneberger2015u} and SegFormer \cite{xie2021segformer}, by training on the $\ourname$ dataset.
According to our experiments, the test accuracy of U-Net and SegFormer trained on the non-augmented $\ourname$ dataset is up to 55.48\% and 82.27\%, respectively.  
The dataset and the code for training models and data augmentation are made available on the GitHub repository through: \href{https://github.com/RobustInsight/ConstScene/}{https://github.com/RobustInsight/ConstScene/}.

\section{Related Work}
\label{sec:related_work}

To the best of our knowledge, $\ourname$ is the first high-resolution public semantic segmentation dataset captured from a construction environment considering a wide range of objects and environmental conditions. 
This study also sheds light on the impact of data augmentation on the segmentation model's effectiveness in handling perturbed scenes.

\textbf{On-Road Vision Datasets.} Several research works have introduced vision datasets designed for applications in on-road autonomous driving \cite{geiger2013vision, cordts2016cityscapes, behley2019semantickitti, yu2020bdd100k, caesar2020nuscenes, huang2018apolloscape}. 
While earlier studies have achieved success in collecting extensive datasets with significant variations, they overlooked the inclusion of construction sites.
In other words, they assumed that autonomous machines would not navigate through construction areas. 

\textbf{Off-Road Vision Datasets.} \cite{kolar2018transfer} focused on guardrail detection in construction sites using a VGG16 model trained on a synthetic dataset containing 6,000 images.
\cite{tajeen2014image} collected images of five different construction machines, including loaders, bulldozers, excavators, backhoe diggers, and rollers. 
The MOCS dataset \cite{xuehui2021dataset} gathered more than 40,000 images from various construction sites for the annotation of 13 distinct types of moving objects. Employing pixel segmentation, they annotated the objects and evaluated their performance on various Convolutional Neural Networks (CNN). 
ACID \cite{xiao2021development} collected 10,000 construction machine images for testing various object detection algorithms. 
SODA dataset \cite{duan2022soda} is a comprehensive object detection dataset for construction sites. SODA was released with more than 20k images encompassing four object categories including worker, material, machine, and layout.

Table~\ref{tab:related_work:off-road_dataset} compares $\ourname$ with off-road autonomous driving datasets.
Compared to prior studies, $\ourname$ is the first to address semantic segmentation in challenging environments, such as rainy, dusty, and fogy. Plus, we employ data augmentation techniques to robustify the model training against perturbed data. Notably, our dataset is openly accessible, ensuring result reproducibility. This facilitates researchers to focus on introducing novel detection techniques rather than repeating the time-consuming data collection process.

\begin{table*}[htbp]
\caption{Summary of off-road autonomous driving datasets. All methods utilized a monocular RGB camera for data collection.} 
\resizebox{\textwidth}{!}{
\begin{tabular}{c|c|c|c|c|c|c}
\hline
\multirow{2}{*}{\textbf{Dataset}}  & \multirow{2}{*}{\textbf{Year}}  & \multirow{2}{*}{\textbf{Task}} & \textbf{Static \& Moving}  & \textbf{Challenging} & \textbf{Data} & \multirow{2}{*}{\textbf{Availability}}  \\

\textbf{} & \textbf{} &  & \textbf{Objects} \textbf{} & \textbf{Weather} & \textbf{Perturbation} & \textbf{}  \\ \hline

Kolar et al. \cite{kolar2018transfer}  & 2018   &  Object Detection & \xmark & \xmark & \xmark & \xmark\\ 

MOCS \cite{xuehui2021dataset} & 2021 &  Semantic Segmentation & \xmark  & \cmark & \xmark & \xmark \\ 

ACID \cite{xiao2021development} & 2021   & Object Detection &  \xmark & \cmark & \xmark & \xmark \\ 

SODA \cite{duan2022soda} & 2022 & Object Detection & \cmark & \cmark & \xmark & \xmark \\ 

\hline
$\ourname$ (Ours) & 2024 & Semantic Segmentation & \cmark &  \cmark & \cmark & \cmark\\ 
\hline
\end{tabular}
}
\label{tab:related_work:off-road_dataset}
\end{table*}

\section{Data Collection and Preparation}
\label{sec:data_collection}

\subsection{Data Collection and Sensor Setup}
\label{sec:data_collection:sensorsetup}

The selection of objects in image datasets is vital for training models to precisely recognize elements such as human, machines, and piles, ultimately enhancing the reliability and decision-making capabilities of self-driving machines in diverse and dynamic construction environments.
In this paper, we try to include commonly used objects in construction environments, including construction machine (wheel-loader), crusher, human, pile, road, and background regions.

The image data collection process poses several challenges, to include variability in the dataset such as lighting and weather conditions, diverse backgrounds, occlusions, and the need to include authentic situations such as existing dust on the camera lens or workers climbing the ladder of a heavy machine.
To address these challenges, we carefully have selected representative samples considering diverse environment conditions to ensure model generalization. 
Plus, we employ data augmentation (Section~\ref{sec:data_collection:augmentation}), to artificially manipulate the input images to simulate variations in weather conditions. 
Occlusion challenges are mitigated by collecting images from multiple angles, while meticulous labeling and annotation processes are undertaken to provide the necessary ground truth for training robust models.
For collecting image data, we utilized an RGB camera featuring a 1280$\times$720 resolution, capturing images in JPEG format at a rate of 15 Frame-Per-Second. 

\subsection{Data Annotation}
\label{sec:data_collection:annotation}

We leverage Roboflow \cite{roboflow2022} tool to annotate images. 
The tool allows us to label objects with semantic segmentation masks, enabling precise annotations for object classes.
For each image in the dataset, a single Portable Network Graphics (PNG) mask is generated to show the segmented regions corresponding to different classes, providing a representation of the pixel-level annotation.
After labeling the whole dataset, images were randomly divided into three categories train, validation, and test.

\subsection{Data Augmentation}
\label{sec:data_collection:augmentation}

Image augmentation plays a crucial role in enhancing the robustness and generalization capabilities of semantic segmentation models \cite{alomar2023data}.
By diversifying the training dataset through changes in contrast, and/or adding perturbations using noise, ultimately we improve the model's ability to perform well on unseen or slightly different data during testing.
In this paper, we leverage three data augmentation techniques by (i) blurring images through taking 25 neighbor pixels and averaging them; (ii) adding pepper-and-salt noise to 1\% of pixels; and (iii) removing eight random regions with the size of 3\% of image size through a cut-out process for imitating dirty and muddy camera lens.
 In the first two methods the added perturbation may not be visible to human eye but still can make the model to miss-classify.
These augmentation techniques have been applied to the images of the train set in the dataset.
Fig.~\ref{fig:augmented_image_threefirst} shows several samples from the dataset where these three perturbation techniques are applied. 

\begin{table}[htbp]
\resizebox{\textwidth}{!}{
\begin{tabular}{c|c|c|c}
\includegraphics[width =\columnwidth]{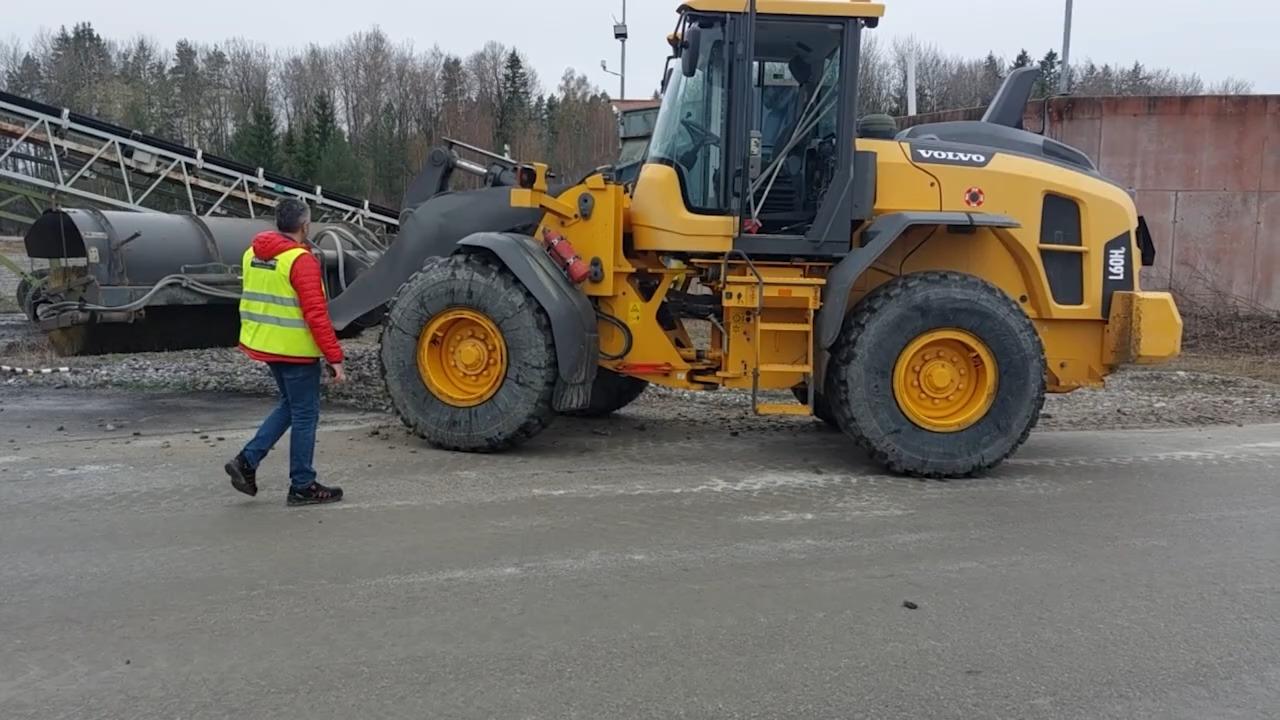}
&
\includegraphics[width =\columnwidth]{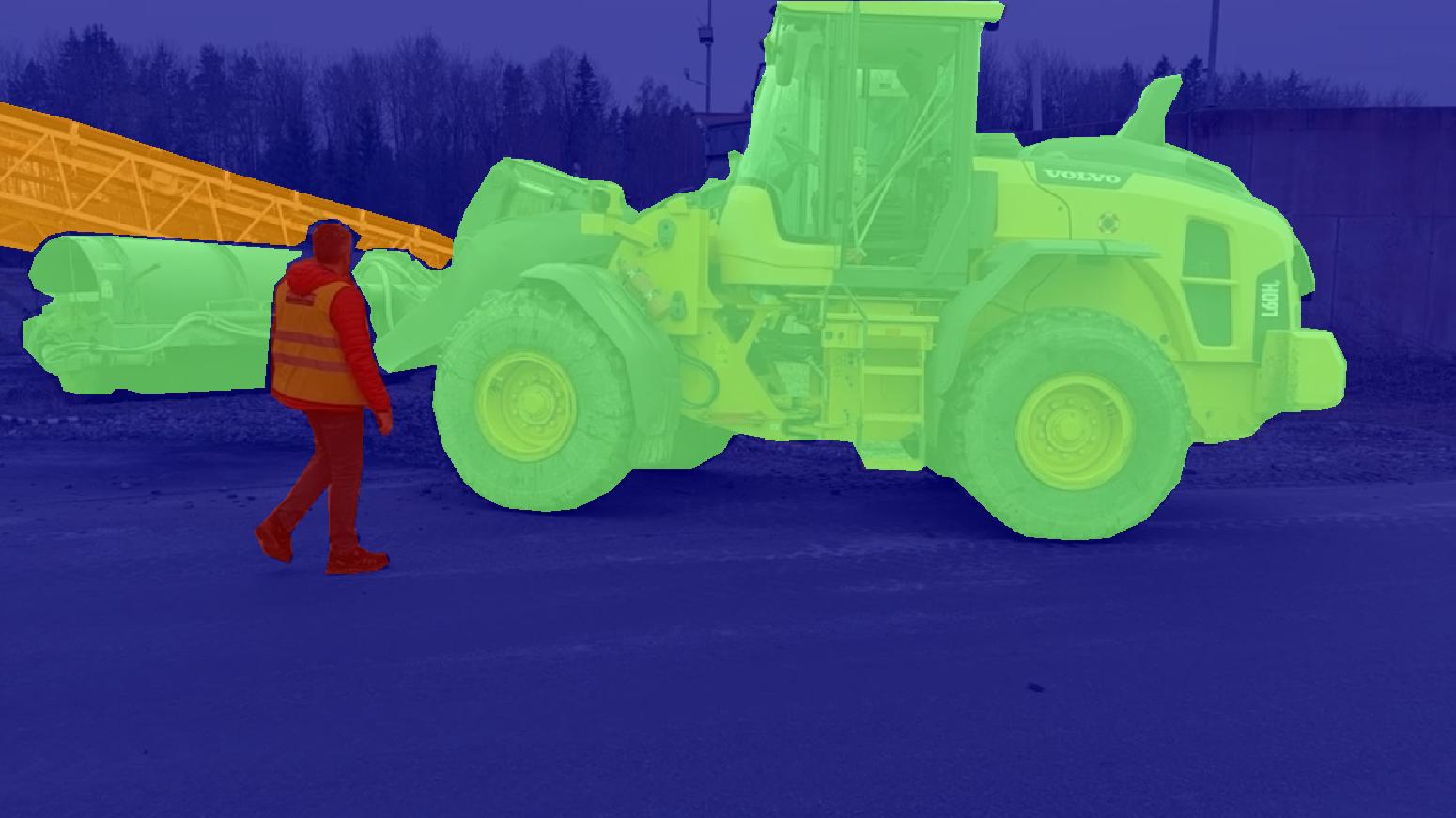}
&
\includegraphics[width =\columnwidth]{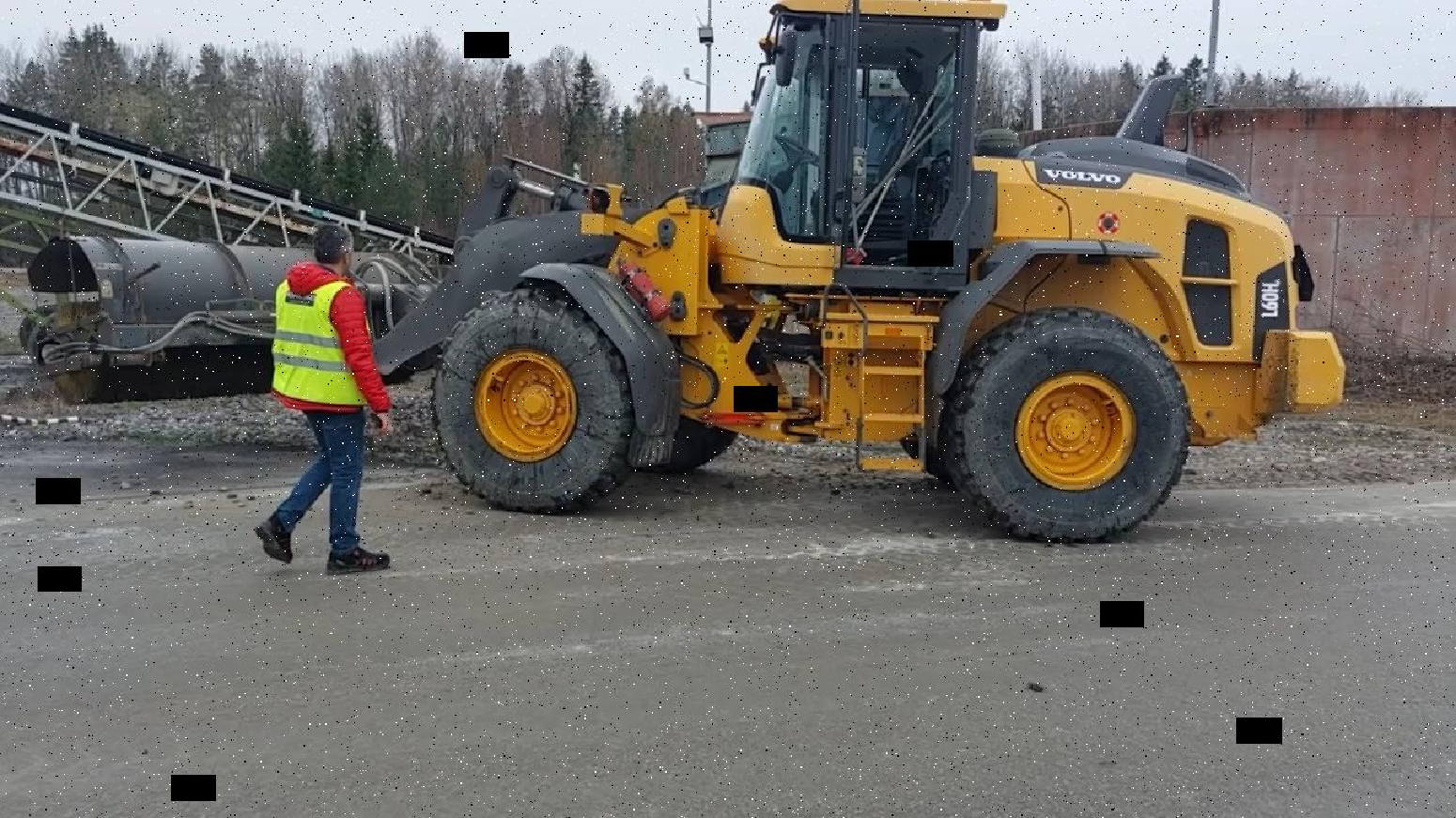}
&
\includegraphics[width =\columnwidth]{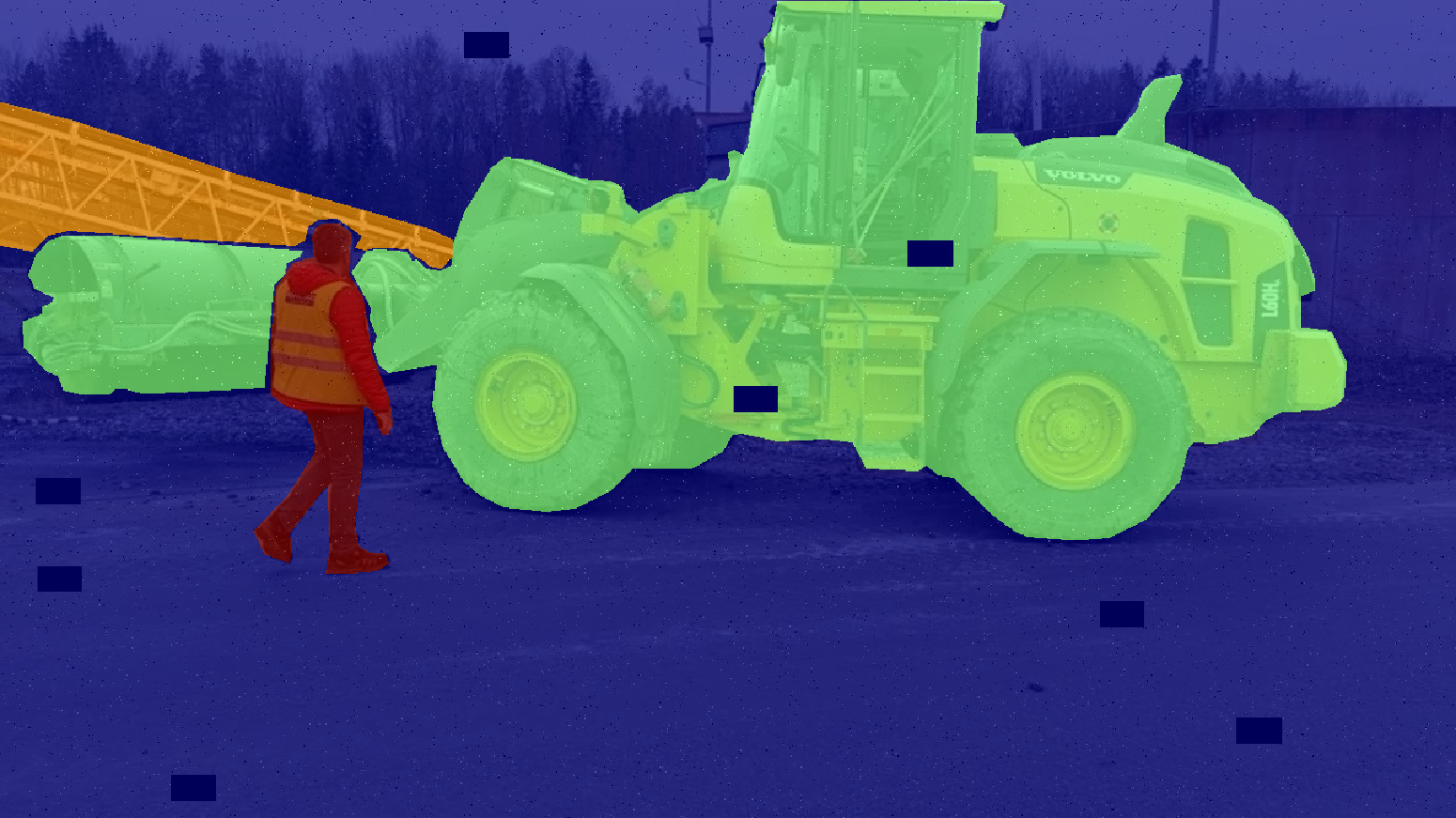}
\\ 
\includegraphics[width =\columnwidth]{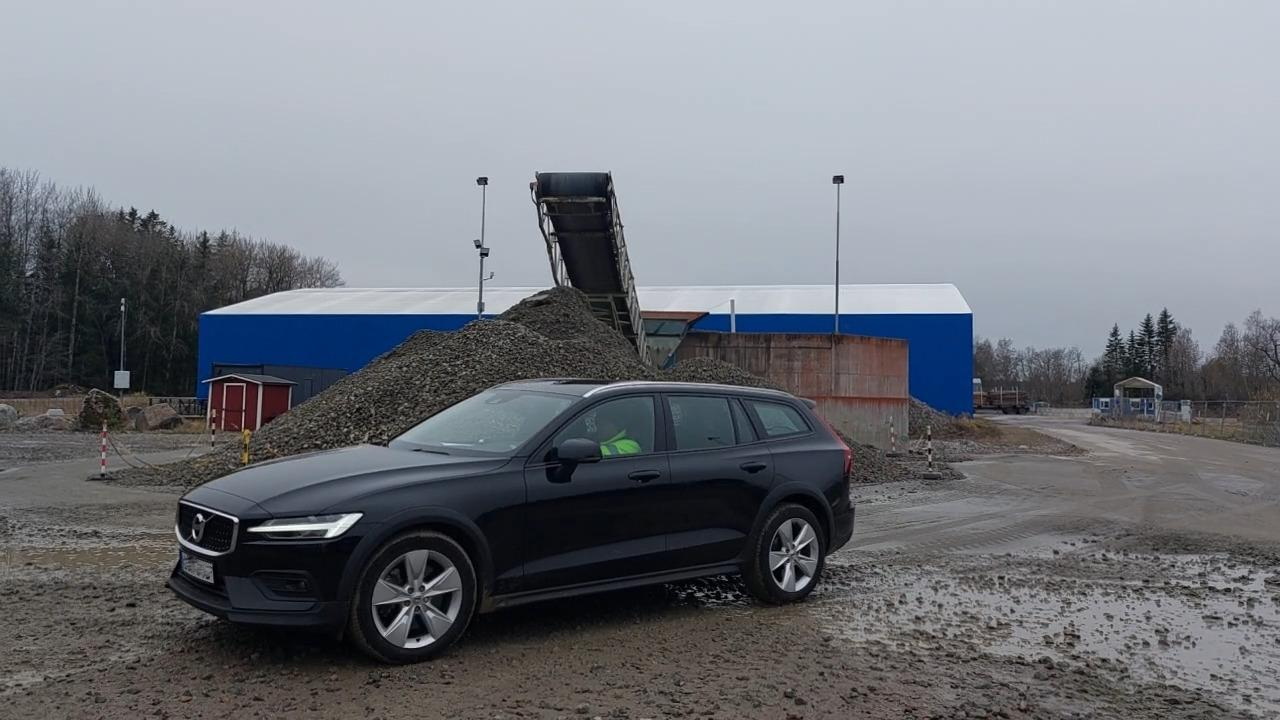}
&
\includegraphics[width =\columnwidth]{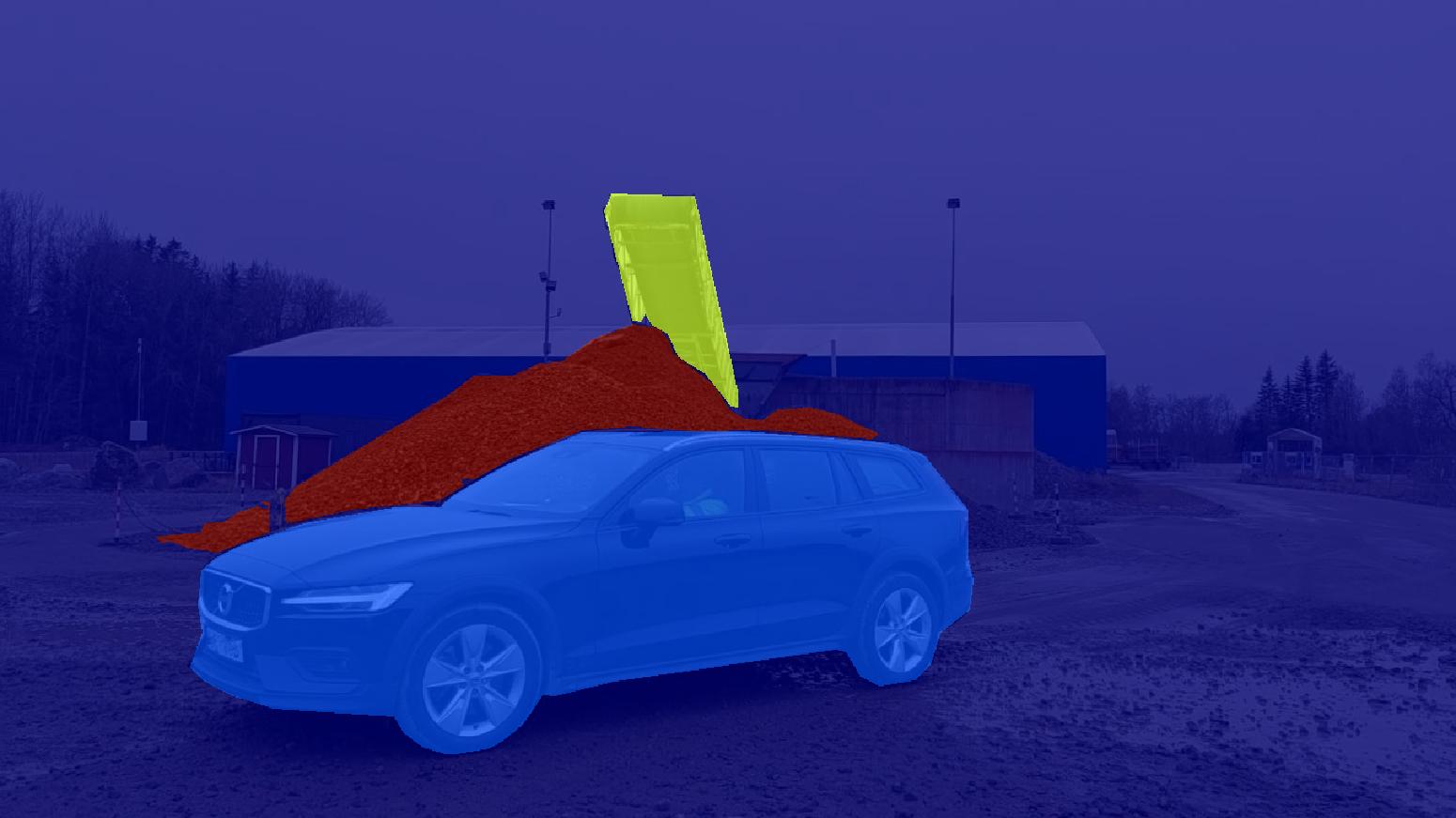}
&
\includegraphics[width =\columnwidth]{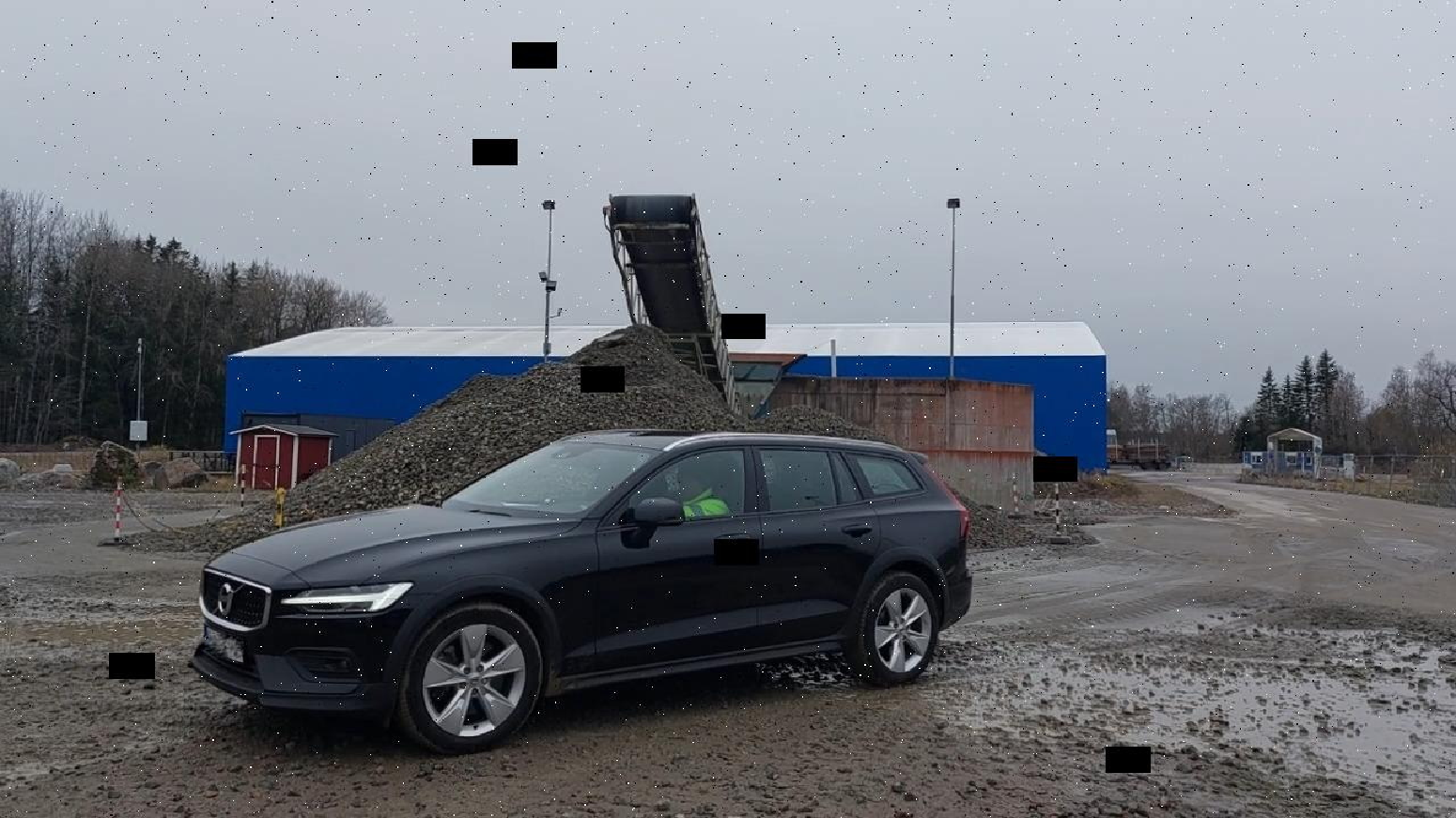}
&
\includegraphics[width =\columnwidth]{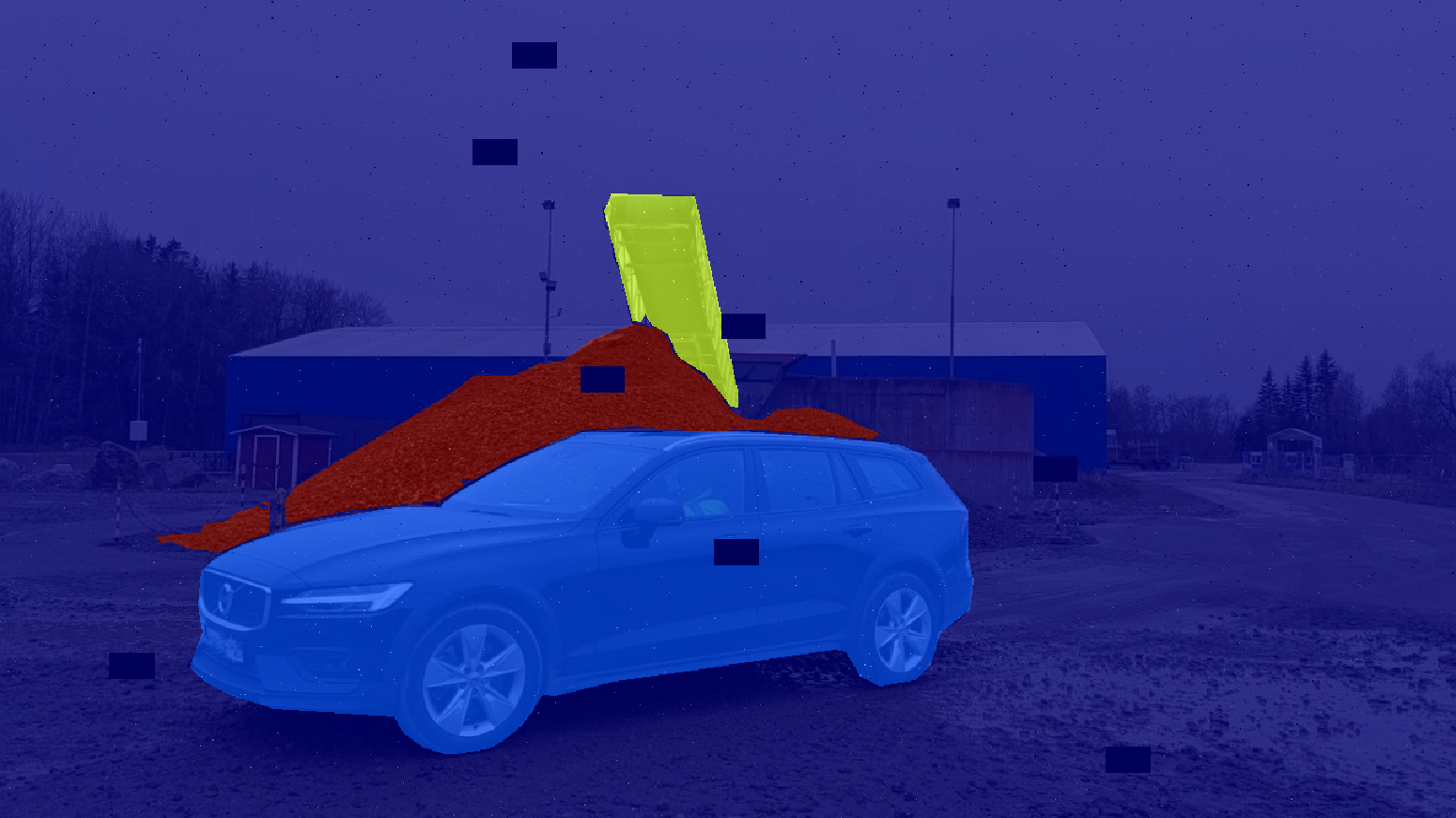}
\\ 
\includegraphics[width =\columnwidth]{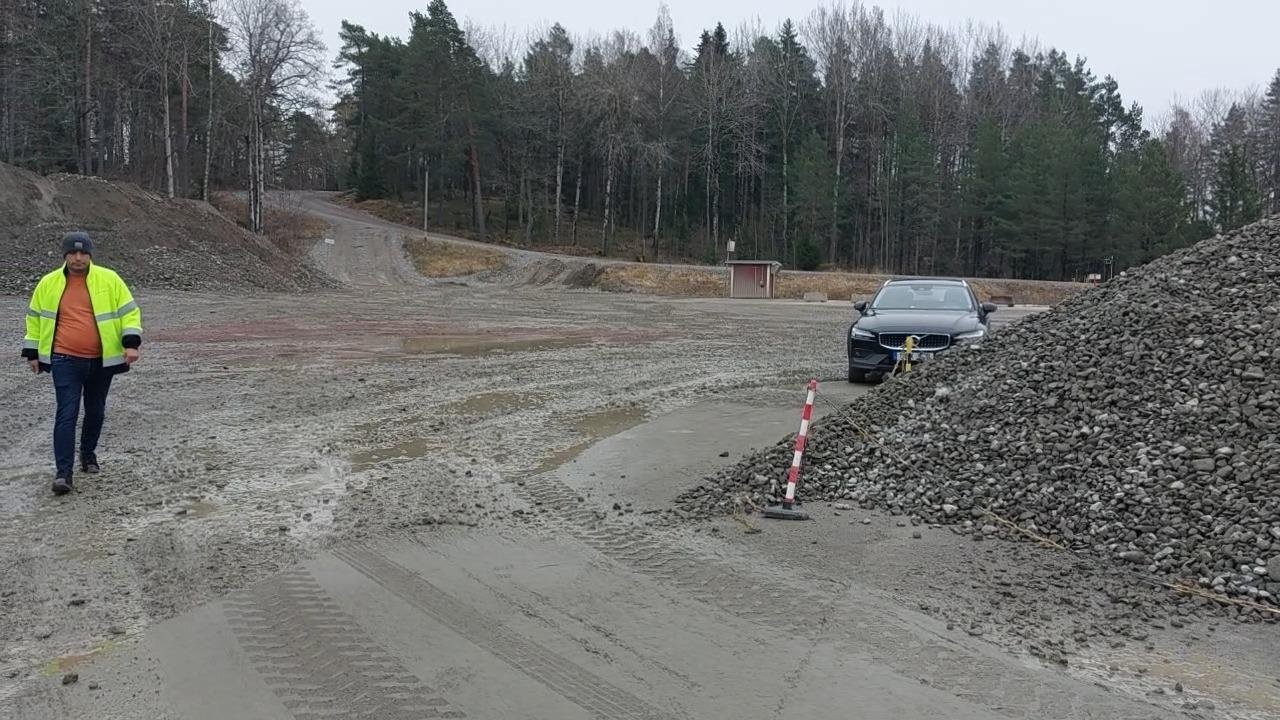}
&
\includegraphics[width =\columnwidth]{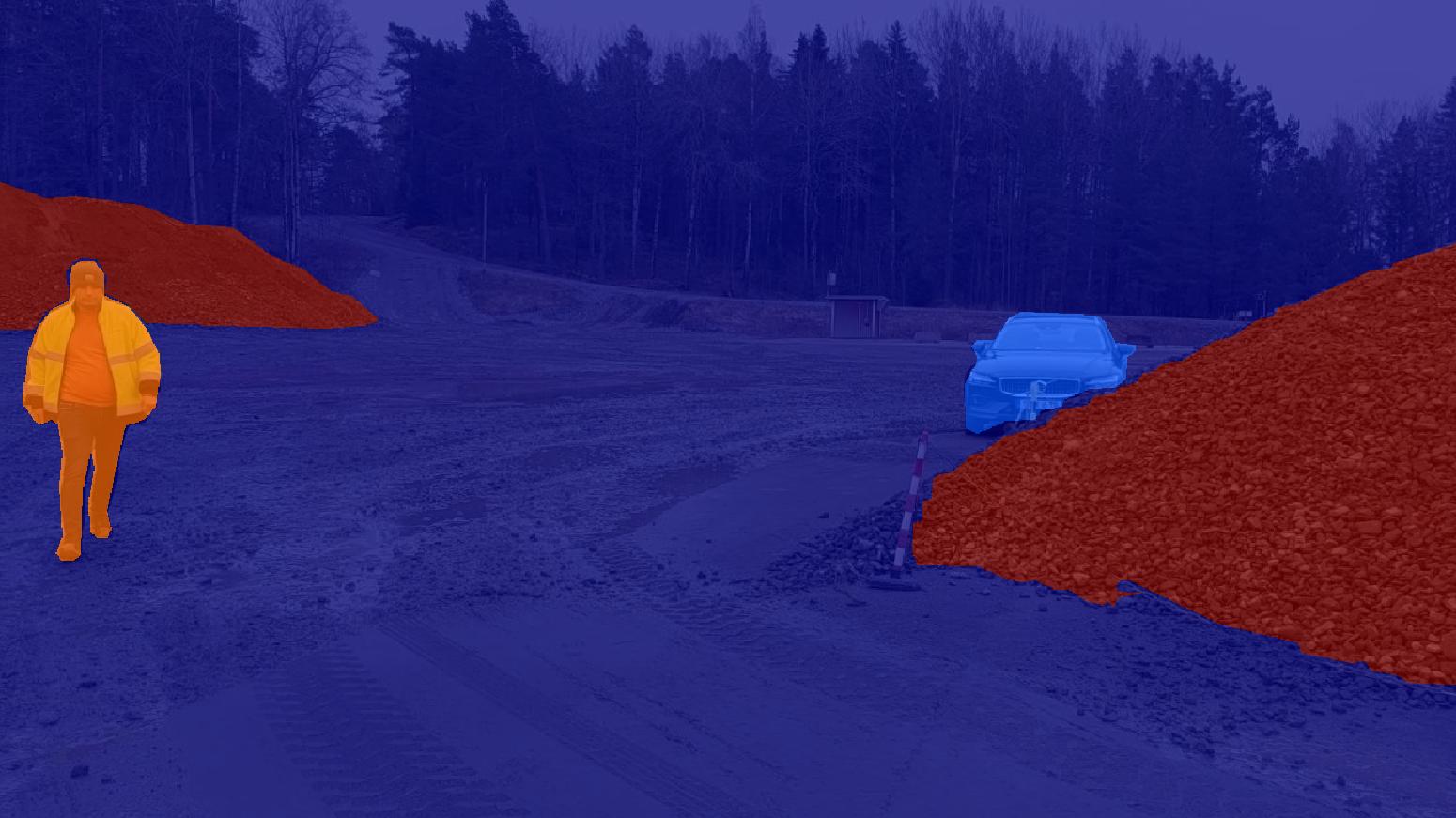}
&
\includegraphics[width =\columnwidth]{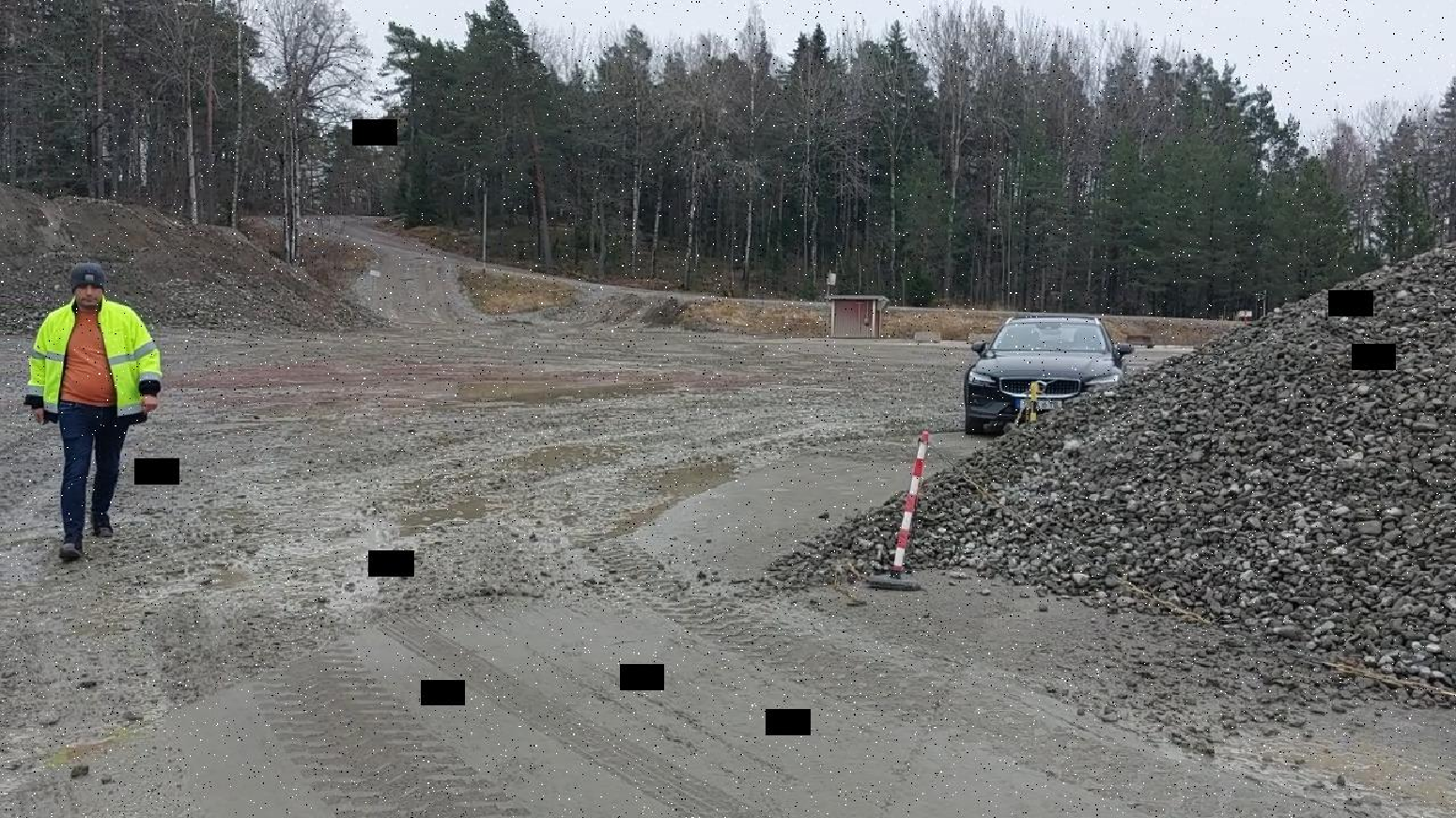}
&
\includegraphics[width =\columnwidth]{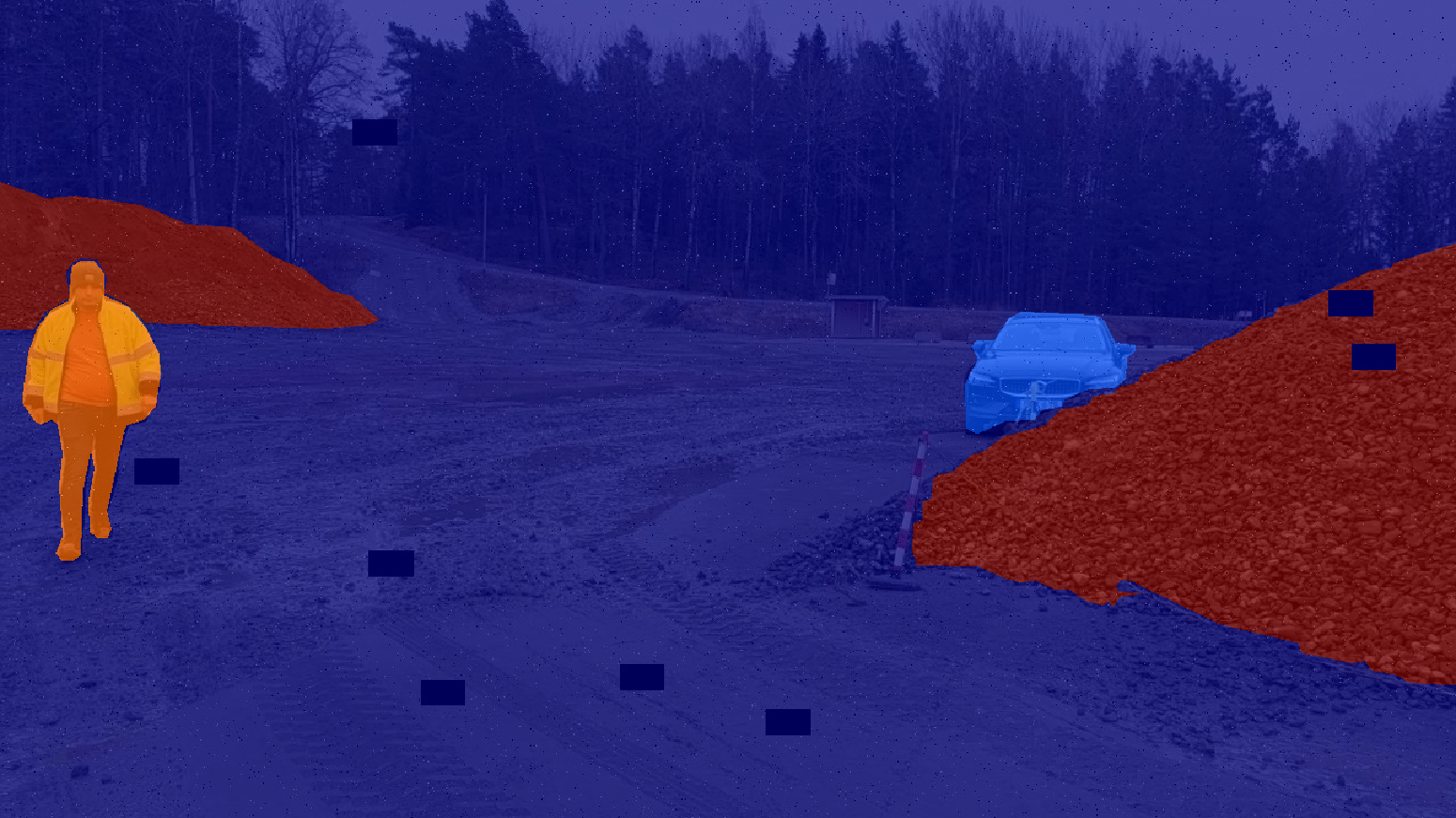}
\\
\textbf{\Huge (a)} & \textbf{\Huge (b)} & \textbf{\Huge (c)} & \textbf{\Huge (d)} \\
\end{tabular}
} 
\captionof{figure}{(a) Original image. (b) Overlay mask of the original image. (c) Augmented image. (d) Overlay mask of the augmented image.}
\label{fig:augmented_image_threefirst}
\end{table}

\subsection{Depth Estimation}
\label{sec:data_collection:depth}

Depth estimation is the task of determining the distance from the camera to objects in a scene \cite{loni2020densedisp, loni2021faststereonet, ming2021deep}. 
Depth estimation is crucial for various applications, such as robotics, autonomous machines, and augmented/virtual reality where environment perception is required for decision making.
In autonomous machines application, depth estimation plays a crucial role for obstacle avoidance that contributes to safety.
Given the significance of depth estimation in autonomous driving, we provide depth maps of images in the dataset using a monocular depth estimation method \cite{ranftl2020towards}. 
Fig.~\ref{fig:samples} shows multiple RGB samples of $\ourname$ with corresponding depth maps across diverse weather and working conditions.
The red circles in Fig.~\ref{fig:samples}.c to Fig.~\ref{fig:samples}.e mark the areas of disparity with low-confidence matches for conditions with the presence of fog, dust, and dirt on the camera lens.
As can be seen, the impact of adverse weather conditions significantly diminishes the accuracy of depth estimation, emphasizing the crucial need to address this challenge in computer vision pipelines employed in construction environments.

\begin{table}[htbp]
\centering
\begin{center}
\resizebox{\linewidth}{!}{
\begin{tabular}{cc}

\includegraphics[width =0.9\columnwidth]{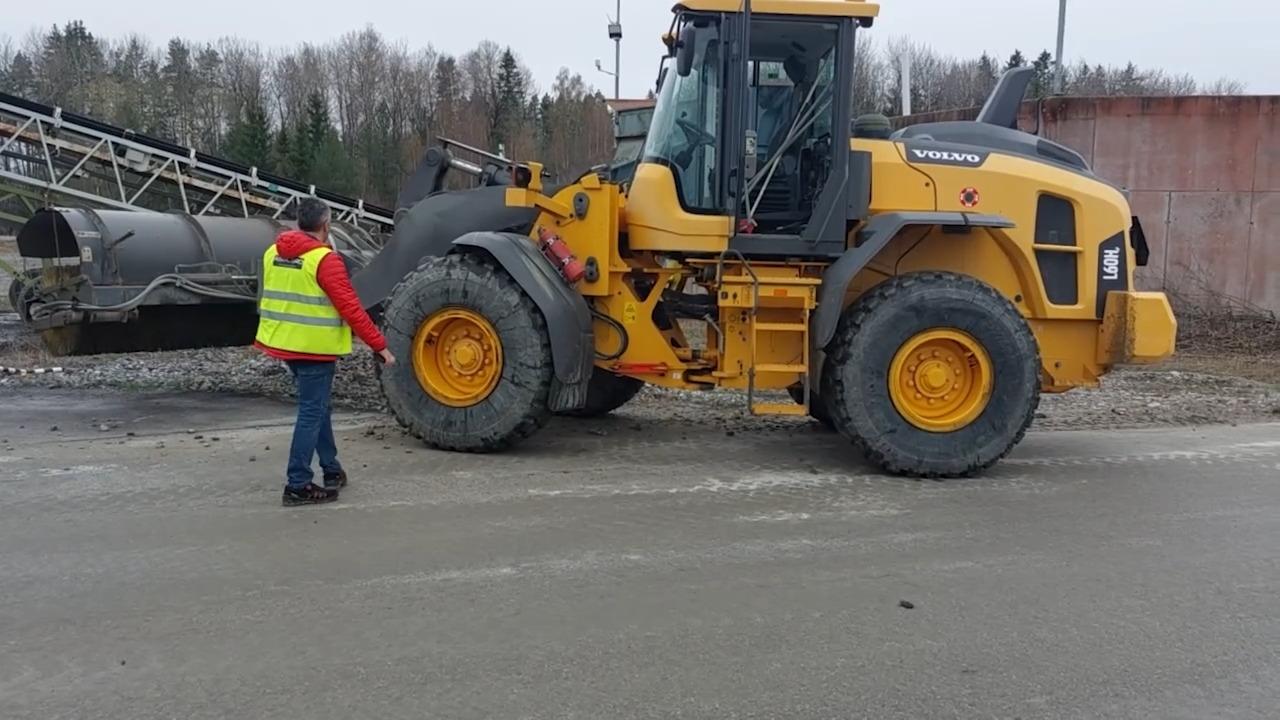}
&
\includegraphics[width =0.9\columnwidth]{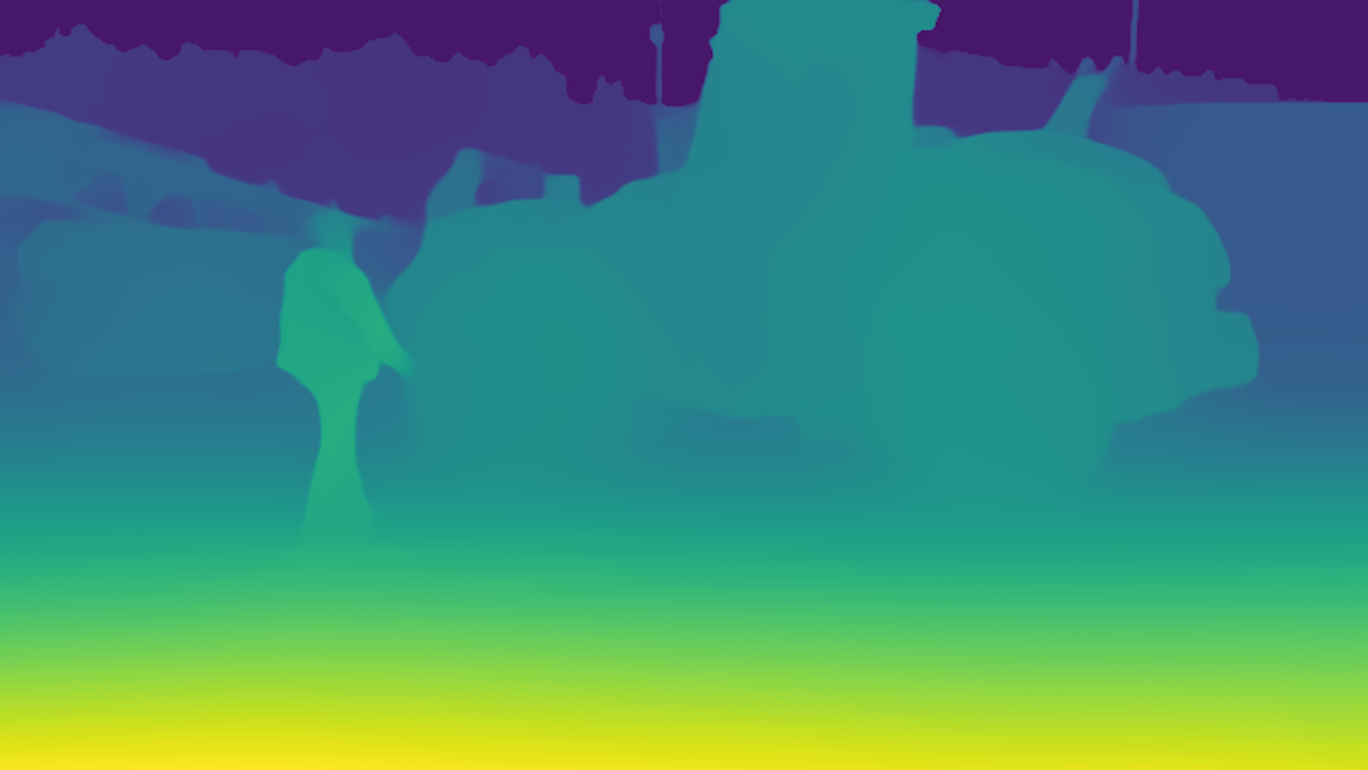}
\\
\multicolumn{2}{c}{ \textbf{\huge (a) Sunny} } \\

\includegraphics[width =0.9\columnwidth]{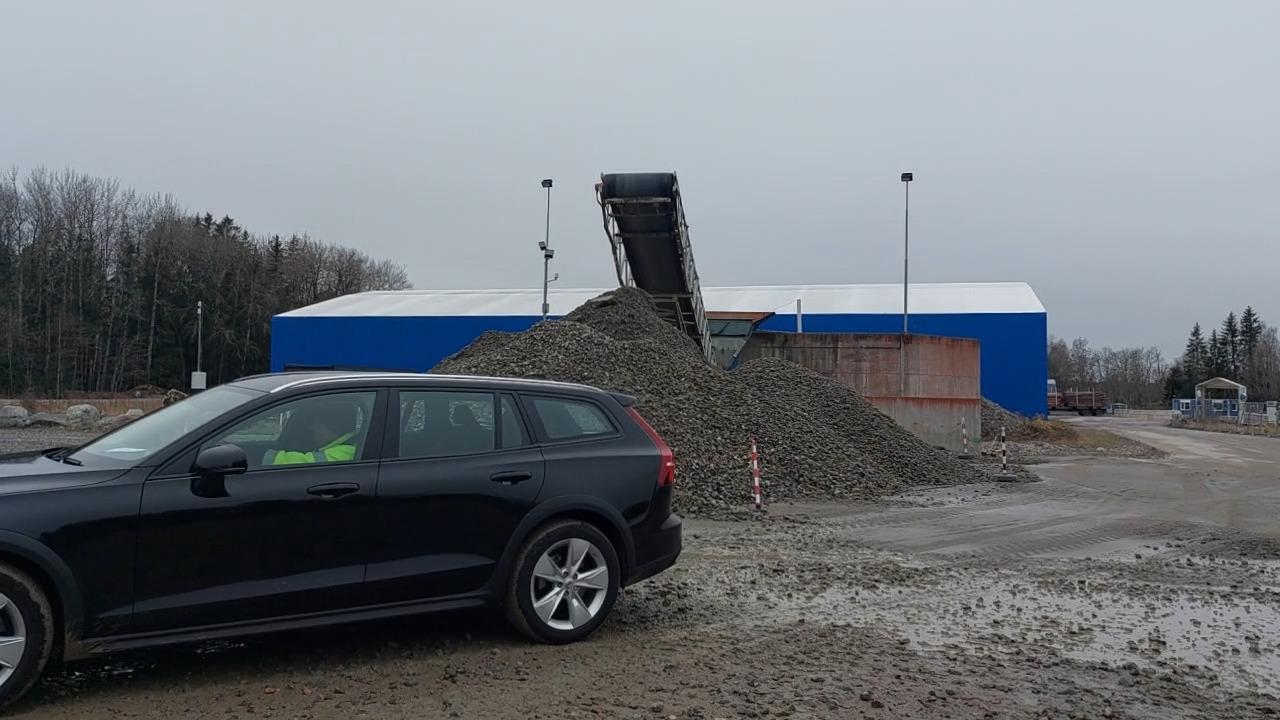}
&
\includegraphics[width =0.9\columnwidth]{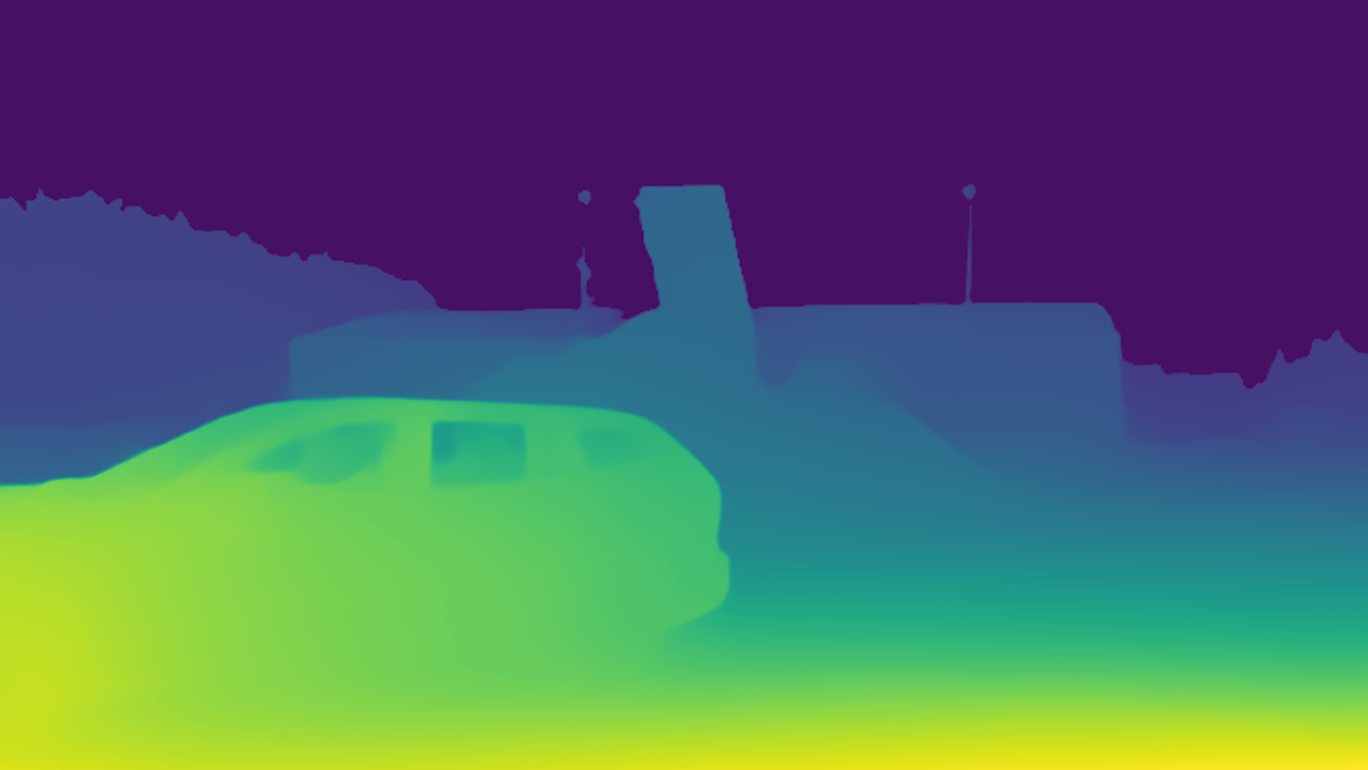}
\\
\multicolumn{2}{c}{ \textbf{\huge (b) Rainy} } \\

\includegraphics[width =0.9\columnwidth]{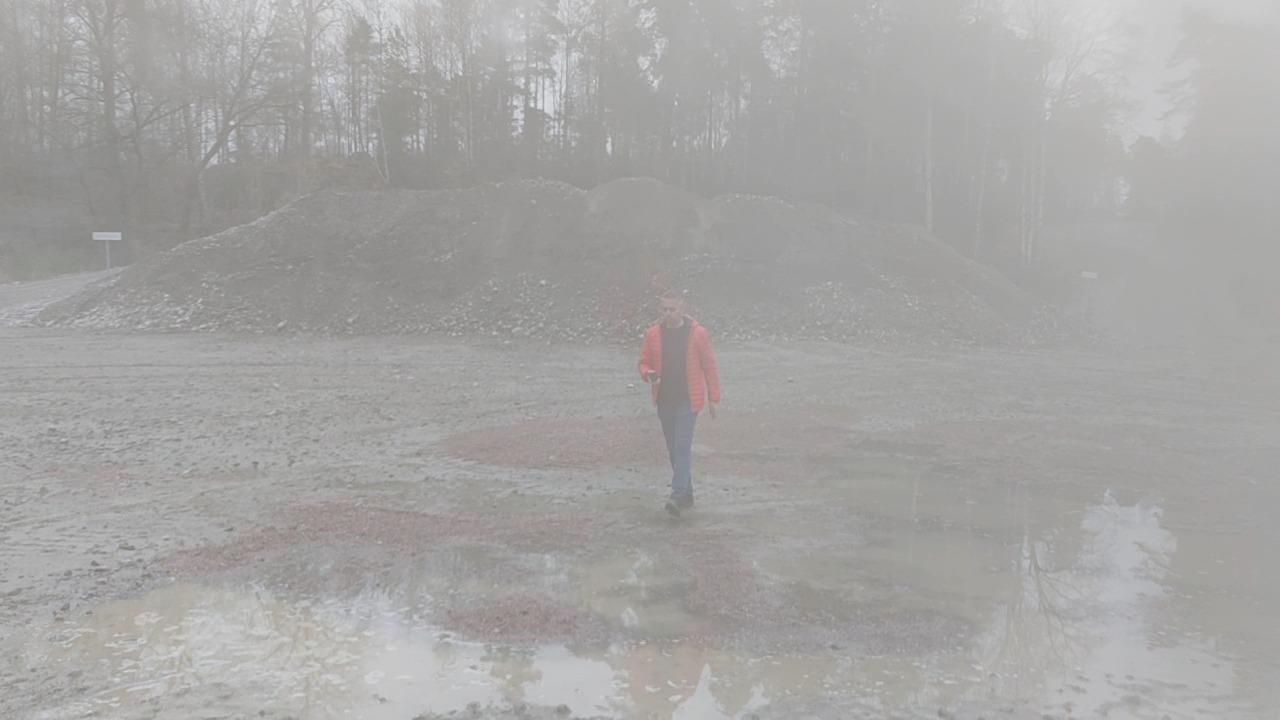}
&
\includegraphics[width =0.9\columnwidth]{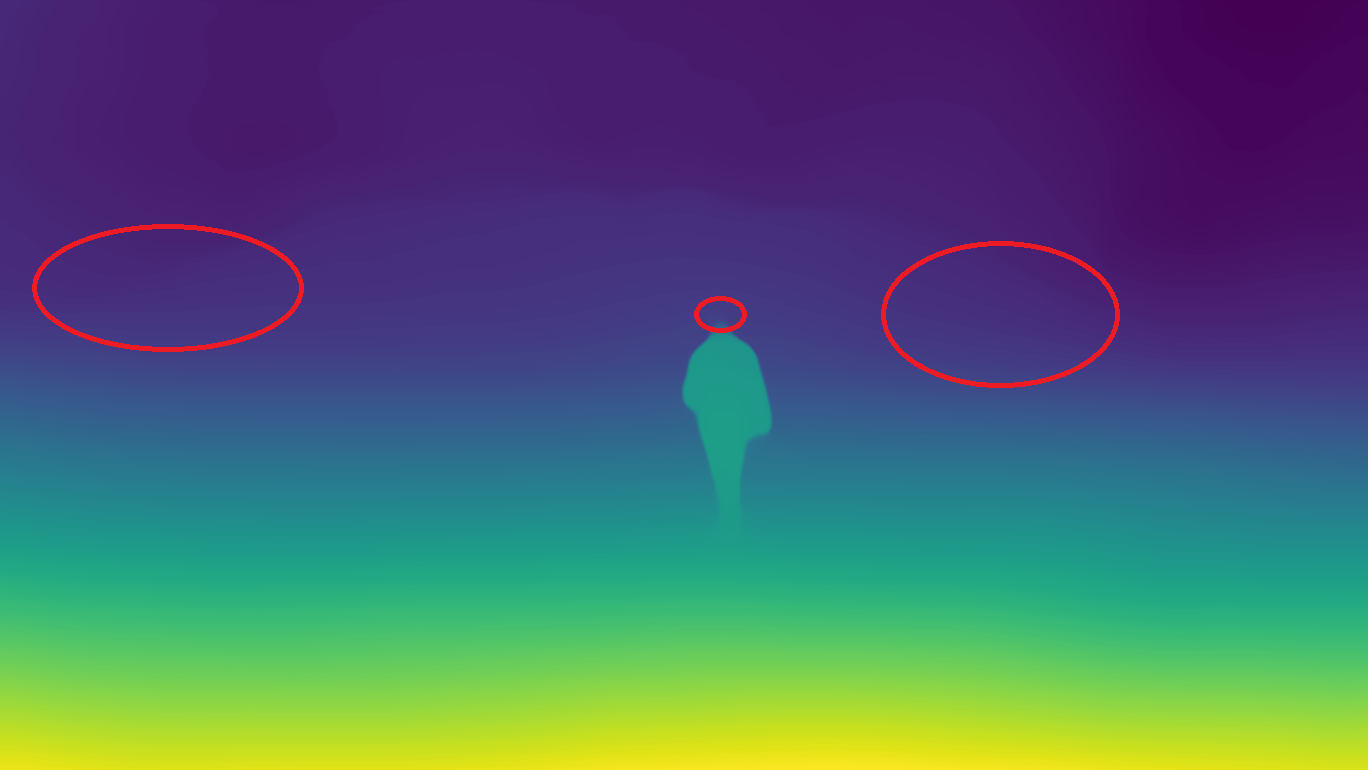}
\\
\multicolumn{2}{c}{ \textbf{\huge (c) Foggy} } \\

\includegraphics[width =0.9\columnwidth]{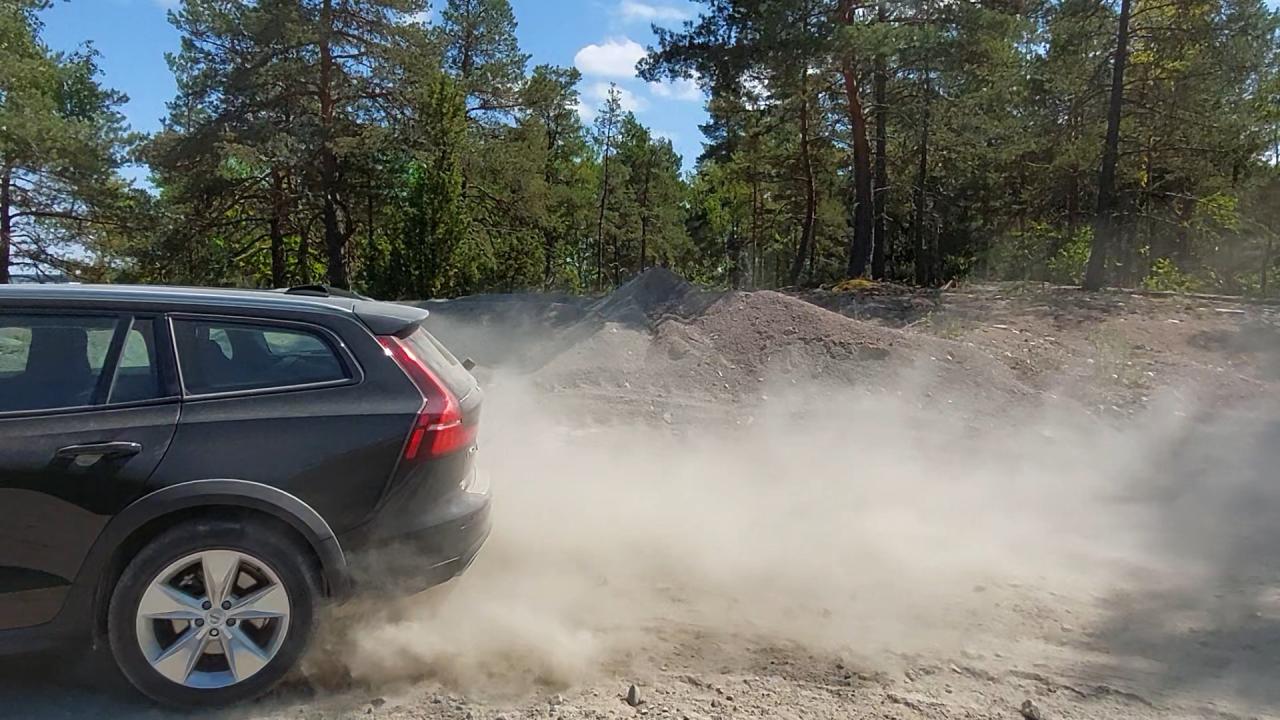}
&
\includegraphics[width =0.9\columnwidth]{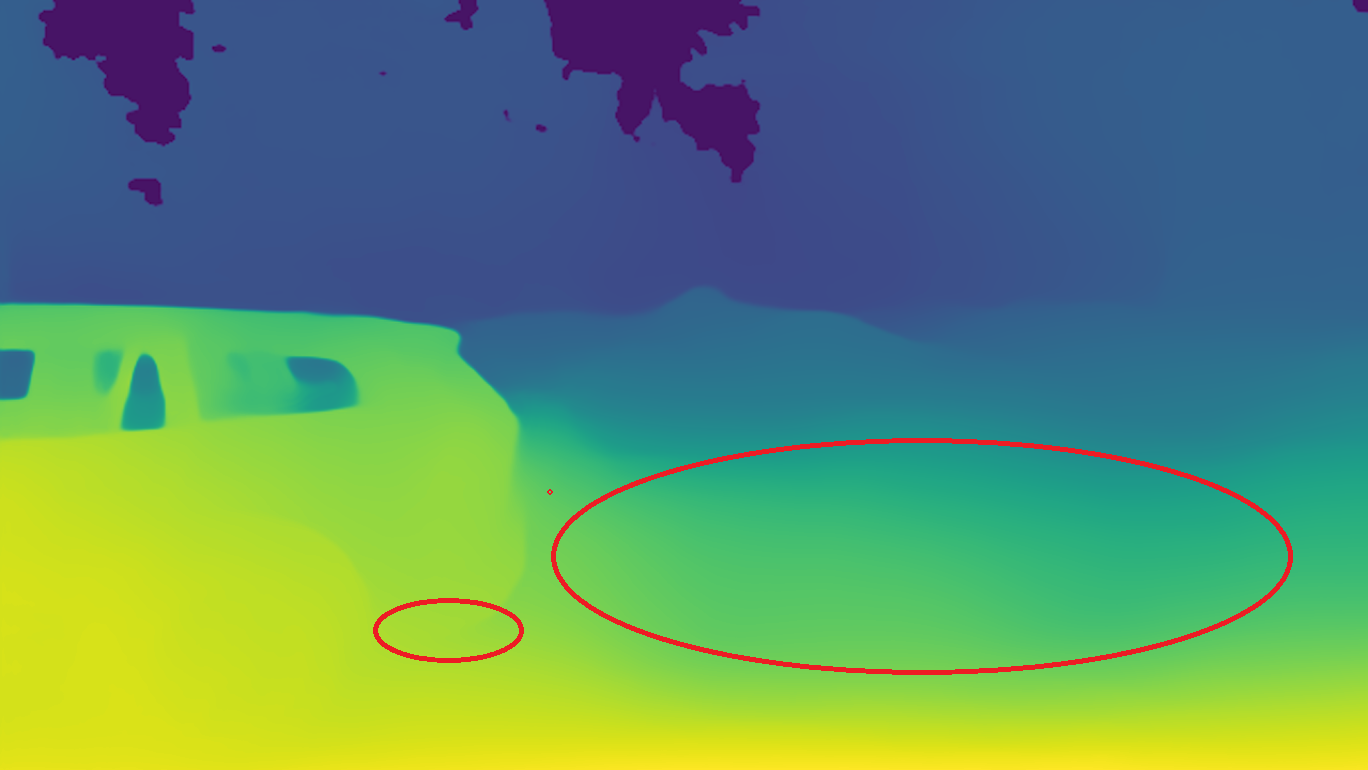}
\\
\multicolumn{2}{c}{ \textbf{\huge (d) Dusty} } \\

\includegraphics[width =0.9\columnwidth]{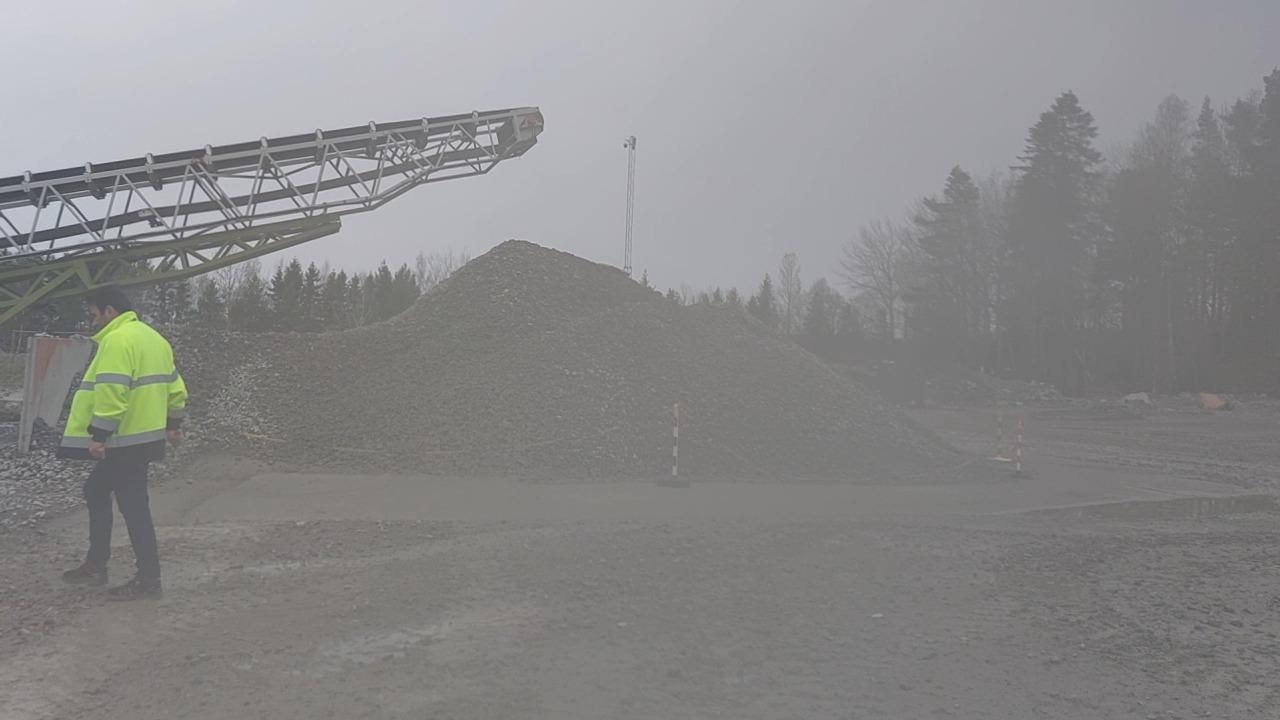}
&
\includegraphics[width =0.9\columnwidth]{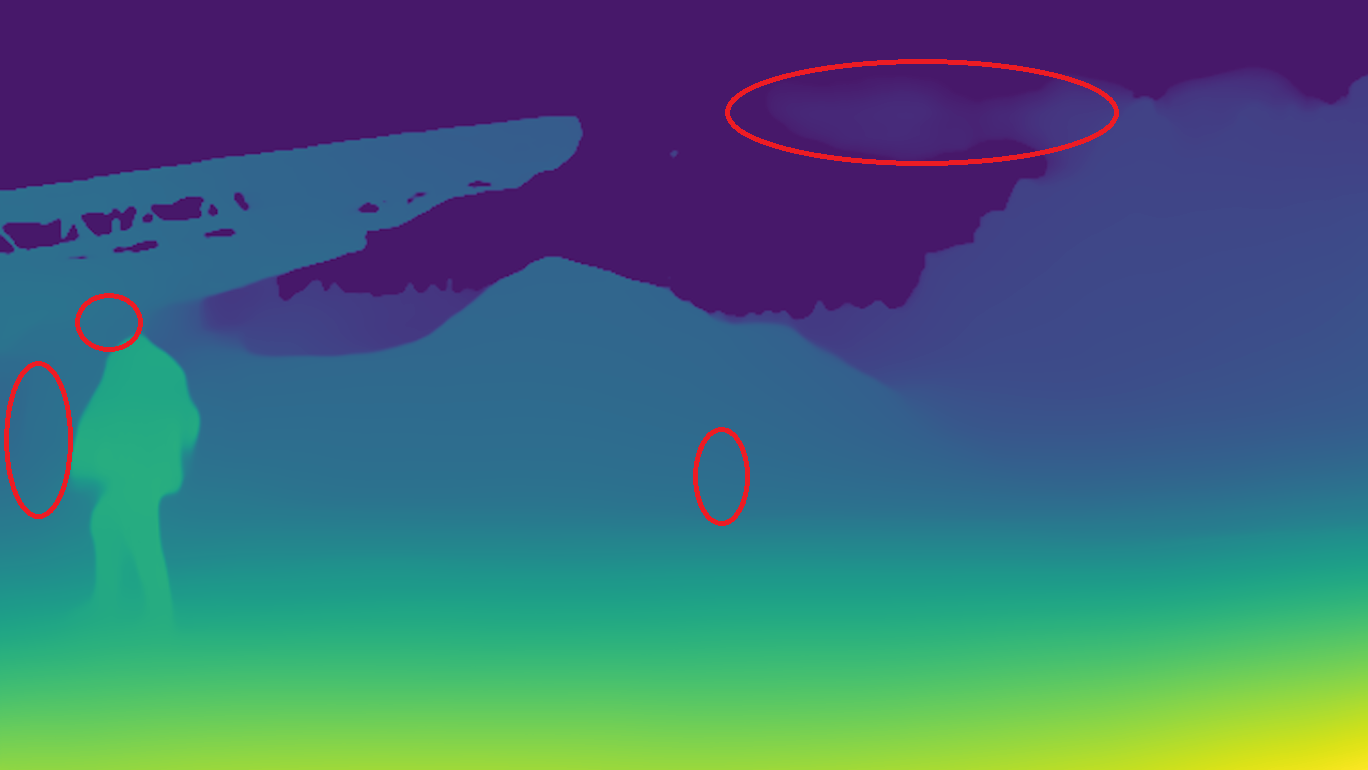}
\\
\multicolumn{2}{c}{ \textbf{\huge (e) Dirty Lens} } \\

\end{tabular}}  
\end{center}
\captionof{figure}{Visual representation of samples from our dataset under various weather conditions, accompanied by the corresponding disparity map. The red circles indicate the disparity region with low-confidence matches.}
\label{fig:samples}
\end{table}

\section{Dataset Statistics}
\label{sec:dataset_statistics}

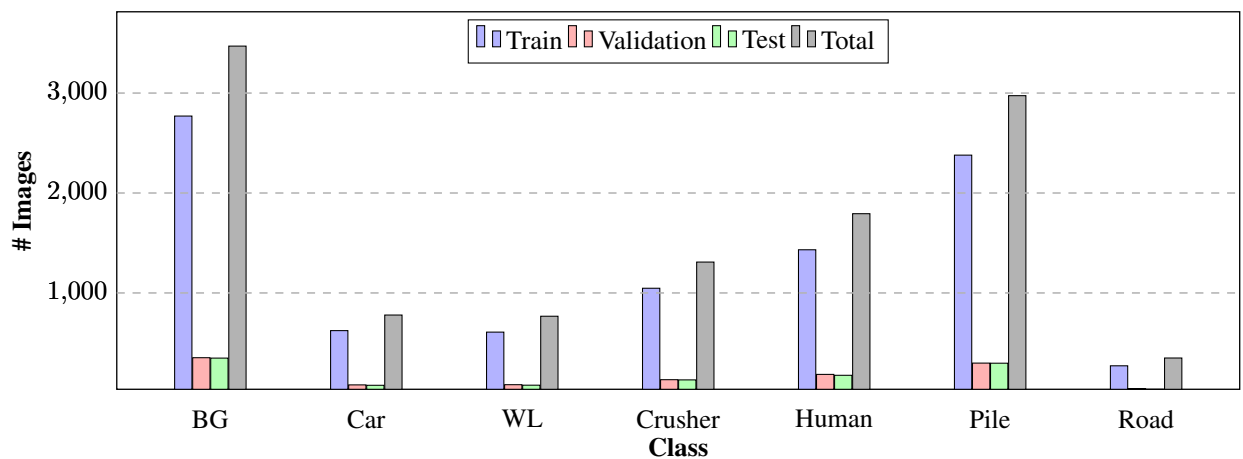
\begin{figure}[htbp]
\centering
\begin{tikzpicture}
\begin{axis}[
    height=0.4\columnwidth,
    width=\textwidth, 
    ybar=0.01cm,
    bar width=0.23cm, 
    enlargelimits=0.10,
    enlarge y limits={value=0.10,upper}, 
    axis on top, 
    axis line style={draw=none}, 
    tick style={draw=none}, 
    axis x line*=bottom, 
    axis y line*=left, 
    axis background/.style={ draw=black, line width=0.5pt},
    legend style={at={(0.5,0.98)},
      anchor=north,legend columns=-1},
    ylabel={\textbf{\# Images}},
    xlabel={\textbf{Class}},
    xticklabel style={text width=1.7cm, align= center}, 
    symbolic x coords={BG, Car, WL, Crusher, Human, Pile, Road},
    xtick=data,
    ymajorgrids,
    every axis plot/.append style={fill},
    extra y ticks={0,1000,2000,3000}, 
    grid style={dashed,gray!60},
    ]
\addplot[fill=blue!30] coordinates {(BG, 2770) (Car, 623) (WL, 608) (Crusher, 1047) (Human, 1432) (Pile, 2379) (Road, 271)};
\addplot[fill=red!30] coordinates {(BG, 352) (Car, 80) (WL, 82) (Crusher, 132) (Human, 185) (Pile, 298) (Road, 42)};
\addplot[fill=green!30] coordinates {(BG, 348) (Car, 76) (WL, 77) (Crusher, 130) (Human, 176) (Pile, 297) (Road, 36)};
\addplot[fill=black!30] coordinates {(BG, 3470) (Car, 779) (WL, 767) (Crusher, 1309) (Human, 1793) (Pile, 2974) (Road, 349)};
\legend{Train, Validation, Test, Total}

\end{axis}
\end{tikzpicture}
\caption{Distribution of images in each class, including background (BG), Car, wheel-loader (WL), crusher, human, pile, and road.}
\label{fig:dataset_distribution}
\end{figure}

$\ourname$ is available in two distinct versions: the original and the augmented version.
The original version comprises 3470 images as training, validation, and test sets, while the augmented version consists of 6240 images.
The partition ratio of the training, validation, and test sets for the original dataset are 80\%, 10\%, and 10\%, respectively.
On the other hand, for the augmented version, these sets were allocated at proportions of 90\%, 5\%, and 5\%.
Fig.~\ref{fig:dataset_distribution} illustrates the class distribution in the dataset, indicating the number of images per class across the training, validation, and test sets in the original dataset.

\section{Experimental Results}
\label{sec:evaluation}

\subsection{Evaluation Metrics} 
\label{sec:evaluation:metrics}

We use mean intersection over union (mIoU) as the evaluation metric for assessing predictions in our experiments.
The Mean Intersection over Union (mIoU) used in semantic segmentation is defined as $mIoU = \frac{1}{N} \sum_{i=1}^{N} \frac{|X_i \cap Y_i|}{|X_i \cup Y_i|}$, where $N$, $X$, and $Y$ are the total number of classes, the set of pixels predicted to belong to class $i$, and the set of pixels in the ground truth that belong to class $i$, respectively.  

\subsection{Experimental Setup} 
\label{sec:evaluation:setup}

\textbf{Training Dataset.} We conduct evaluations on two versions of $\ourname$: (i) $D1$, representing the original dataset without any data augmentation; and (ii) $D2$, denoting the original dataset incorporating the first three augmentation techniques ($blurring + pepper\mhyphen and\mhyphen salt~noise + regions~removal$).

\textbf{Model Details.} For evaluation, we leverage U-Net model \cite{ronneberger2015u} with ResNet18 and ResNet50 \cite{he2016deep} encoder architectures; and SegFormer \cite{xie2021segformer} as a state-of-the-art Vision Transformer (ViT) \cite{dosovitskiy2020image} model.
A U-Net is a CNN architecture designed for image segmentation tasks.
It features a U-shaped structure, comprising a contracting path for capturing high-level features and a corresponding expansive path for precise segmentation.
Skip connections maintain spatial details.
The U-Net effectively balances local and global contexts, making it versatile for various computer vision applications.
In our exploration of ViT, we examined various iterations of the advanced SegFormer model \cite{xie2021segformer}.
SegFormer has two main modules: a hierarchical transformer encoder; and a lightweight multi-layer-perception decoder to predict the final segmentation mask.

\textbf{Training Details.} Table~\ref{tab:training} summarizes training parameters for U-Net and SegFormer models.
We utilize NVIDIA\textsuperscript{\textregistered} A5000 for training U-Net and SegFormer models on $\ourname$.
Considering sustainability, it's noteworthy to highlight the carbon emissions associated with the training process.
By utilizing the machine learning CO$_2$ impact calculator tool \footnote{https://mlco2.github.io/impact/}, the total carbon footprint for all experiments is roughly 8.65 kg CO$_2$.

\begin{table}[htbp]
\centering
\caption{Summary of training parameters for U-Net and SegFormer models.}
\label{tab:training}
\resizebox{0.6\textwidth}{!}{%
\begin{tabular}{c|c|c}
\hline
\textbf{Parameter}    & \textbf{U-Net \cite{ronneberger2015u}} & \textbf{SegFormer \cite{xie2021segformer}} \\ \hline
Optimizer             &  Adam   &  Adam  \\
Initial Learning Rate &  3$\times$10$^{-4}$ & 2$\times$10$^{-5}$ \\
\# Epochs             &  50  &  64  \\
Batch Size            & \{1000, 1500\}  &  \{200, 500\}  \\

\hline
\end{tabular}%
}
\end{table}

\subsection{Prediction Results} 
\label{sec:evaluation:results}

\textbf{Analyzing Results of U-Net.} Table~\ref{tab:unet-performance} presents the test results of U-Net with various backbone architectures trained on the original $\ourname$ with and without the incorporation of data augmentation techniques.
Results show that U-Net with the ResNet50 encoder trained on the non-augmented dataset ($D1$) provides the best results (55.48\%) compared to other training settings.
In addition, data augmentation ($D2$) provides up to 11.59\% higher accuracy for U-Net with the ResNet50 encoder demonstrating that the model excels at capturing the diverse patterns found in real-world.

\begin{table}[htbp]
\centering

\caption{Performance of U-Net \cite{ronneberger2015u} with different backbone architectures trained on $\ourname$.}
\label{tab:unet-performance}
\resizebox{\textwidth}{!}{%
\begin{tabular}{c|c|c|c|c|c}
\hline
 \textbf{Encoder}  & \textbf{\# Parameters}  & \textbf{Dataset}  &  {\textbf{Batch Size}} & {\textbf{Train Time (Hour)}} &  \textbf{mIoU (\%)}  \\
\hline
  ResNet18 & 13.7M & $D1$   & 1500 & 1 & 51.93  \\
  ResNet18 & 13.7M & $D2$  & 1500 & 1.7  & 55.76  \\
  \hdashline
  ResNet50 & 31M & $D1$  & 1000 & 2.5 &  55.48   \\
  ResNet50 & 31M & $D2$  & 1000 & 3 & 67.07    \\
\hline
\end{tabular}}
\end{table}

\textbf{Analyzing Results of SegFormer.} Table~\ref{tab:segformer-performance} shows the test results of SegFormer trained on $\ourname$.
The findings reveal that SegFormer-B5, when trained on the original dataset ($D1$), achieves an accuracy of 82.27\%, surpassing SegFormer-B0's results by an improvement of 35.3\%.

\textbf{Analyzing Results of Data Augmentation.} When SegFormer-B5 undergoes training using the dataset enhanced with data augmentation $D2$, it yields superior accuracy compared to the non-augmented dataset.
This suggests that the $blurring + pepper\mhyphen and\mhyphen salt~noise + regions~removal$ augmentation techniques contribute to the model's improved ability to learn variations present in real-world data.

\begin{table}[htbp]
\centering
\caption{Performance of SegFormer \cite{xie2021segformer} with different model sizes trained on $\ourname$.}
\label{tab:segformer-performance}
\resizebox{\textwidth}{!}{%
\begin{tabular}{c|c|c|c|c|c}
\hline
\textbf{SegFormer-$*$} & \textbf{\# Parameters} &  {\textbf{Dataset}} &  {\textbf{Batch Size}} & {\textbf{Train Time (Hour)}} &  {\textbf{mIoU (\%)}} \\
\hline
B0 & 3.7M & $D1$  & 500 & 3.7 & 46.96  \\
B0 & 3.7M & $D2$  & 500 & 7.5 & 64.84  \\
\hdashline
B5 & 84.6M & $D1$  & 200 & 13 & 82.27 \\
B5 & 84.6M & $D2$  & 200 & 27 & 87.09 \\
\hline
\end{tabular}}
\end{table}

\subsection{Advantages of $\ourname$ in Defending Against Adversarial Attacks} 
\label{sec:evaluation:discussion}

Adversarial attacks \cite{chakraborty2018adversarial, salimi2023learning, salimi2023saraf} involve the deliberate manipulation of input data to cause a model to produce incorrect or misleading outputs.
Adversarial attacks can significantly threaten the reliability and security of machine learning systems. 
Challenging weather conditions and natural phenomena, such as poor lighting and shadow, are categorized as a type of black-box adversarial attack \cite{zhong2022shadows, wang2022survey}.
$\ourname$ can be leveraged to improve the model's robustness against such adversarial attacks through the adversarial training strategy.
By deliberately exposing the model to adverse weather samples during training, (i) the model learns to defend against such attacks and becomes more adept at recognizing and mitigating adversarial inputs; and (ii) the model is better equipped to handle variations and distortions in real-world data, making it more resilient against adversarial inputs.

\section{Conclusion and Future Work}
\label{sec:conclusion}

We presented a novel large-scale dataset, dubbed $\ourname$, for studying semantic object detection in construction environment across various weather and environmental conditions with extensive annotations for comprehensive understanding and analysis.
We have extended our dataset by employing data augmentation techniques aligned with environmental hazards prevalent in construction sites.
We also showed the results of two popular semantic segmentation models trained on our proposed dataset.
In future work, semantic segmentation in construction environments can be further improved by (i) considering objects from other construction segments such as mining; and (ii) generating synthetic data using generative adversarial networks. 
Our new findings and the proposed dataset pose a high potential in stimulating the development of innovative techniques to enhance the efficacy of object detection models against adversarial attacks.

\subsubsection{Acknowledgements} This work was supported by Volvo Construction Equipment AB, and the Swedish research financier for universities, KK-stiftelsen, through the \href{https://sacsys.github.io/main/}{SACSys} project.

\end{document}